\documentclass[twoside,11pt]{article}

\usepackage{jmlr2e}

\usepackage{amsmath,latexsym}
\usepackage{multirow}
\usepackage{bm}
\usepackage{hyperref}
\usepackage{afterpage}
\usepackage{booktabs}
\usepackage{makecell}

\graphicspath{{Figures/}}

\newcommand{\stirling}[2]{S(#1,#2)}
\newcommand{\tableref}[1]{Table \ref{#1}}
\newcommand{\figureref}[1]{Figure \ref{#1}}
\newcommand{\secref}[1]{Section \ref{#1}}
\newcommand{\appendixref}[1]{Appendix \ref{#1}}
\newcommand{\equationref}[1]{Equation \ref{#1}}

\firstpageno{1}

\begin{document}

\title{The Impact of Random Models on Clustering Similarity}

\author{\name Alexander J. Gates \email ajgates@indiana.edu
\AND
   \name Yong-Yeol Ahn \email yyahn@indiana.edu \\
   \addr Department of Informatics and Program in Cognitive Science\\
   Indiana University\\
   919 East 10th Street \\
   Bloomington, IN 47408, USA
   }

\editor{}

\maketitle

\begin{abstract}%
Clustering is a central approach for unsupervised learning.
After clustering is applied, the most fundamental analysis is to quantitatively compare clusterings.
Such comparisons are crucial for the evaluation of clustering methods as well as other tasks such as consensus clustering.
It is often argued that, in order to establish a baseline, clustering similarity should be assessed in the context of a random ensemble of clusterings.
The prevailing assumption for the random clustering ensemble is the permutation model in which the number and sizes of clusters are fixed.
However, this assumption does not necessarily hold in practice; for example, multiple runs of K-means clustering returns clusterings with a fixed number of clusters, while the cluster size distribution varies greatly.
%
Here, we derive corrected variants of two clustering similarity measures (the Rand index and Mutual Information) in the context of two random clustering ensembles in which the number and sizes of clusters vary.
In addition, we study the impact of one-sided comparisons in the scenario with a reference clustering.  
The consequences of different random models are illustrated using synthetic examples, handwriting recognition, and gene expression data.
We demonstrate that the choice of random model can have a drastic impact on the ranking of similar clustering pairs, and the evaluation of a clustering method with respect to a random baseline; thus, the choice of random clustering model should be carefully justified.

\end{abstract}

\begin{keywords}
clustering comparison, clustering evaluation, adjustment for chance, Rand index, normalized mutual information
\end{keywords}

\section{Introduction}
Clustering is one of the most fundamental techniques of unsupervised learning and one of the most common ways to analyze data.  
Naturally, numerous methods have been developed and studied (\citealp{Jain2010dataclustering}).
To interpret clustering results, it is crucial to compare them to each other.
For instance, the evaluation of a clustering method is usually carried out by comparing the method's results with a planted reference clustering, assuming that the more similar the method's solution is to the reference clustering, the better the method.
This is particularly common in the field of complex networks in which clustering similarity measures are used to justify the performance of community detection methods (\citealp{Danon2005communitycompare}; \citealp{Lancichinetti2009communitycompare}). 
As quantitative comparison is a fundamental operation, it plays a key role in many other tasks.
For instance, comparisons of clusterings can facilitate taxonomies for clustering solutions, can be used as a criteria for parameter estimation, and form the basis of consensus clustering methods (\citealp{Meila2005axiomclusteringscompare, Vinh2009correctingnmi, Yeung2001genemodel}).

Among the many clustering comparison methods (see, e.g.\ \citealp{Meila2005axiomclusteringscompare}; \citealp{Pfitzner2009clusteringpairs}), two of the most prominent measures are the Rand index (\citealp{Rand1971randindex}) and the Normalized Mutual Information (NMI, \citealp{Danon2005communitycompare}).
In both cases, the similarity score exists in the range $[0,1]$, where $1$ corresponds to identical clusterings and $0$ implies maximally dissimilar clusterings. 
However, in practice, both measures do not efficiently use the full range of values in between $0$ and $1$, with many comparisons concentrating near the extreme values (\citealp{Vinh2009correctingnmi, Hubert1985adjustedrand}).
This makes it difficult to directly interpret the results of a comparison.

Thus, it is often argued that clustering similarity should be assessed in the context of a random ensemble of clusterings (\citealp{Vinh2009correctingnmi, Hubert1985adjustedrand, Dubien1981extendedrandmodels, Dubien2004randmoments, Albatineh2006correctionchance, Romano2014correctingnmi, Zhang2015nmicompare, Romano2016adjusting}) and rescaled (see \equationref{eq:adjchance}).
Such a correction for chance establishes a baseline by using the expected similarity of all pair-wise comparisons between clusterings specified by a random model; the resulting similarity values have a new interpretation that facilitates comparisons within a set of clusterings.
Specifically, once corrected for chance, a similarity value of $1$ still corresponds to identical clusterings, but a value of $0$ now corresponds to the expected value amongst random clusterings.
Positive values of corrected similarity better reflect an intuitive comparison of clusterings (\citealp{Hubert1985adjustedrand, Steinley2016varianceari}).
The correction may also introduce negative values when two clusterings are less similar than expected by chance.

The correction procedure requires two choices: \emph{a model for random clusterings} and \emph{how clusterings are drawn from the random model}.  
However, even the existence of these choices is usually ignored or relegated to the status of technical trivialities.
Here, we demonstrate that these choices may dramatically affect results, and therefore the choice of a particular model for random clusterings should be justified based on the understanding of the clustering scenario.
%
%
A poor choice of the random model may ``not be random enough'' and encode crucial features of the clusterings in all of the random clusterings, providing a poor baseline.
At the same time, a random model may be ``too random'' in which crucial features are lost in a sea of random clusterings that are not representative of the particular problem.  
Characterizing random models is an important topic of research across statistical physics, network science, and combinatorial mathematics (\citealp{Sethna2006statistical, Goldenberg2010statnetmodels, Mansour2012setpartitions}).
Yet, despite the importance of random model selection, almost no study that uses clustering comparison provides a justification for their choice of random model.

By far, the most common approach to correct clustering similarity for chance assumes that both clusterings are uniformly and independently sampled from the \emph{permutation model} ($M_{\text{perm}}$).
In the permutation model, the number and size of clusters within a clustering are fixed, and all random clusterings are generated by shuffling the elements between the fixed clusters.
However, the premises of the permutation model are frequently violated; in many clustering scenarios, either the number of clusters, the size distribution of those clusters, or both vary drastically (\citealp{Hubert1985adjustedrand, Wallace1983comparehierclusteringscomment}).  
For example, K-means clustering, probably the most common technique, fixes the number of clusters but not the sizes of those clusters (\citealp{Jain2010dataclustering}).  
Later, we explore a real example in which K-means produces clusterings with large variations in the clusterings' cluster size sequences.
%
This suggests that comparing K-means clusterings based on $M_{\text{perm}}$ is misleading.

\begin{table}[b!]
	\centering
	\includegraphics[scale = 0.8]{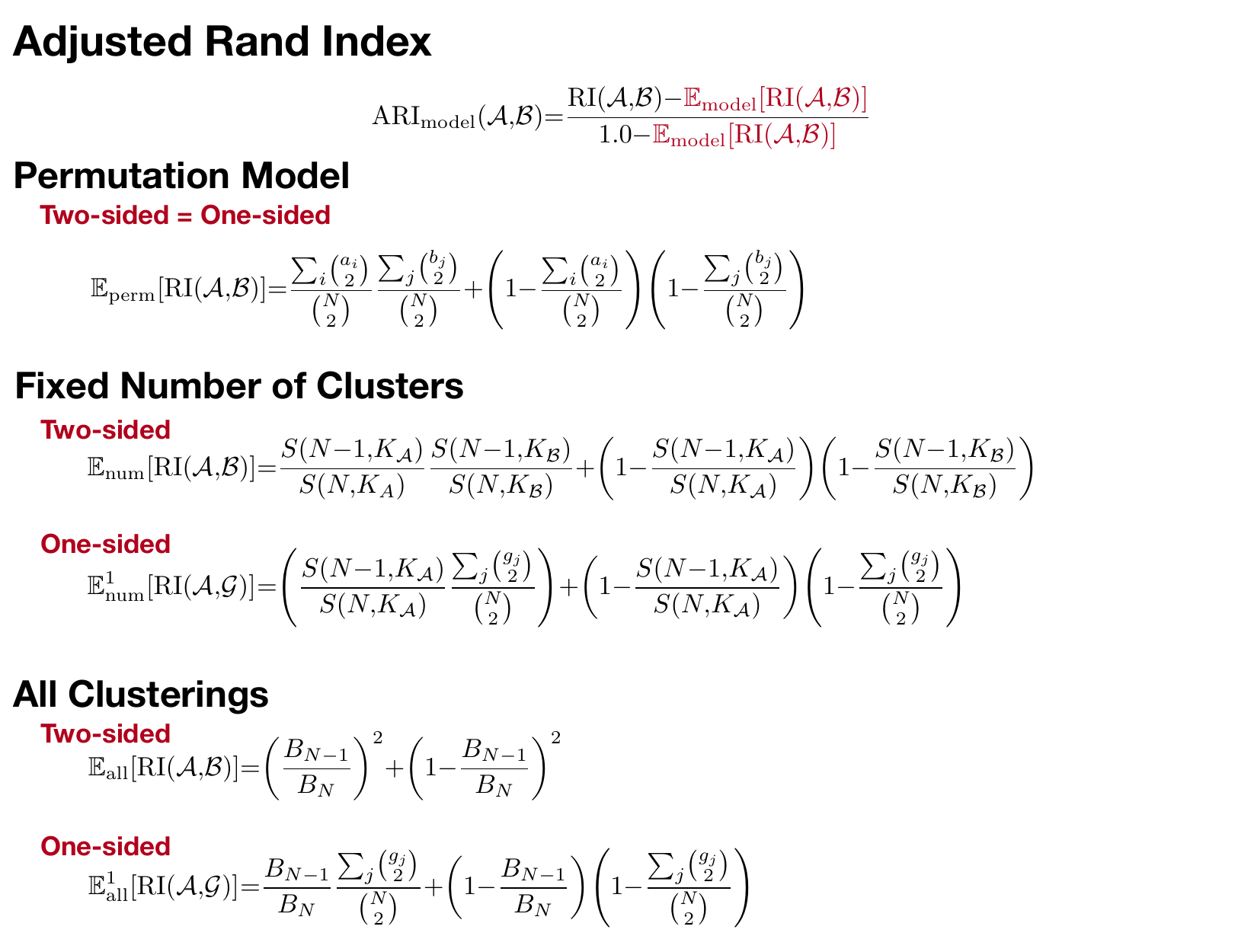}
	\caption{The expected Rand index between two random clusterings $\mathcal{A}$ and $\mathcal{B}$ of $N$ elements, or random clustering $\mathcal{A}$ and reference clustering $\mathcal{G}$ under different random models.  Details and derivations are given in \secref{sec:rand}.}
	\label{tbl:expectedrand}
\end{table}

\afterpage{
\begin{table}[h]
	\centering
	\includegraphics[scale = 0.8]{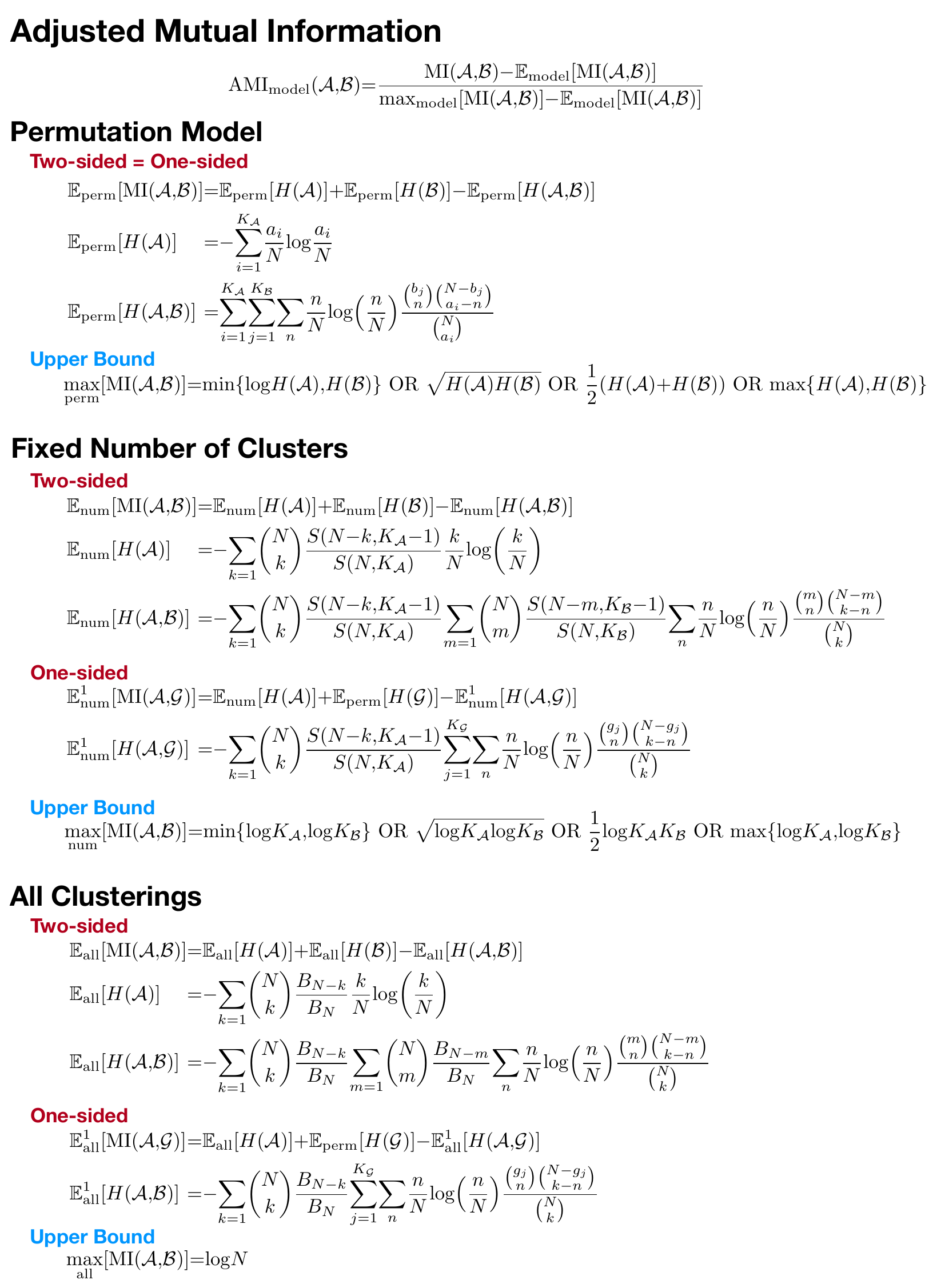}
	\caption{The expected Mutual Information between two random clusterings $\mathcal{A}$ and $\mathcal{B}$ of $N$ elements, or random clustering $\mathcal{A}$ and reference clustering $\mathcal{G}$ under different random models.  Details and derivations are given in \secref{sec:mi}.}
	\label{tbl:expectedmi}
\end{table}
\clearpage }

Furthermore, even the assumption that both clusterings were randomly drawn from the same random model (a two-sided comparison) is often problematic.
For example, when comparing against a given reference clustering, it is more reasonable to find the expected similarity of the reference clustering with all of the random clusterings from the random model.
This \emph{one-sided} comparison accounts for the fixed structure of the reference clustering which is always present in the comparisons, providing a more meaningful baseline.

Here, we present a general framework to adjust measures of clustering similarity for chance by considering a broader class of random clustering models and one-sided comparisons.
Specifically, we consider two other random models for clusterings: a uniform distribution over the ensemble of all clusterings of $N$ elements with the same number of clusters ($M_{\text{num}}$), and a uniform distribution over the ensemble of all clusterings of $N$ elements ($M_{\text{all}}$).
The resulting expectations for the Rand index under all three random models are summarized in \tableref{tbl:expectedrand} and for Mutual Information in \tableref{tbl:expectedmi}, with the full derivations given in \secref{sec:rand} and \secref{sec:mi} respectively.
The adjusted similarity measures used throughout this work rescale the Rand and MI measures by these expectations according to \equationref{eq:adjchance}.
We also introduce one-sided variants of the adjusted Rand index and adjusted Mutual Information when using the $M_{\text{num}}$ or $M_{\text{all}}$ random models (for $M_{\text{perm}}$, the one-sided similarity is equivalent to the two-sided case).

The impact of our framework is illustrated in the case of two common tasks for adjusted clustering similarity measures: 1) ranking the similarity between pairs of clusterings (or finding the most similar clustering pair), and 2) evaluating the performance of a clustering method with respect to a random baseline.
In \secref{sec:results}, these tasks are demonstrated in the context of several examples: a synthetic clustering example, K-means clustering of a handwritten digits data set (MNIST), and an evaluation of hierarchical clustering applied to gene expression data.
Our results demonstrate that both the choice of random model for clusterings and the choice of one-sided comparisons can affect results significantly.
Therefore, we argue that clustering comparisons should be accompanied by a proper justification for the random model.


\section{Clusterings}
We first explicitly introduce a clustering of elements.
Given a set of $N$ distinct elements $V = \{v_{1},\ldots,v_{N}\}$ (i.e. data points or vertices), a clustering is a partition of $V$ into a set $\mathcal{C} = \{C_1,\ldots, C_{K_{\mathcal{C}}}\}$ of $K_{\mathcal{C}}$ non-empty disjoint subsets of $V$, the clusters, $C_k$, such that 
	\begin{enumerate}
		\item $\forall C_i, C_j$ if $i\neq j$, then $C_i\cap C_j = \varnothing$
		\item $\bigcup_{k=1}^{K_{\mathcal{C}}}C_k = V$.
	\end{enumerate}
Each clustering specifies a sequence of cluster sizes, namely, letting $c_i = |C_i|$ be the size of the $i$-th cluster, then the sequence of cluster sizes is $[c_1, c_2,\ldots, c_{K_{\mathcal{C}}}]$.

Throughout this paper, we focus on the similarity of two clusterings over the same set of $N$ labeled elements, $\mathcal{A}=\{A_1,\ldots,A_{K_\mathcal{A}}\}$ (with $K_\mathcal{A}$ clusters of sizes $a_i$) and $\mathcal{B}=\{B_1,\ldots,B_{K_\mathcal{B}}\}$ (with $K_\mathcal{B}$ clusters of sizes $b_j$).

\section{Correction for Chance}
Given a clustering similarity measure $s$ and a random model for clusterings: model, the expected clustering similarity $\mathbb{E}_{\text{model}}[s]$ of pair-wise comparisons within the random ensemble defined by the model corrects $s$ for chance as follows (\citealp{Hubert1985adjustedrand})
\begin{equation}
	\frac{s - \mathbb{E}_{\text{model}}[s]}{s_{\max} - \mathbb{E}_{model}[s]}.
	\label{eq:adjchance}
\end{equation}
The denominator rescales the adjusted similarity by the maximum similarity of pair-wise comparisons within the ensemble $s_{\max}$ so identical clusterings always have a similarity of $1.0$. 
For some clustering similarity measures, the value of $s_{\max}$ is independent of the random model used; for example, the Rand index is always bounded above by $1.0$.  However, in the case of mutual information, the value of $s_{\max}$ depends on the random model used.

\section{Rand Index}
\label{sec:rand}
The Rand index (\citealp{Rand1971randindex}) compares the number of element pairs which are either co-assigned to the same cluster, or assigned to different clusters in both clusterings, to the total number of element pairs.
The most common formulation of the Rand index focuses on the following four sets of the $\binom{N}{2}$ element pairs: 
$N_{11}$ the number of element pairs which are grouped in the same cluster in both clusterings, 
$N_{10}$ the number of element pairs which are grouped in the same cluster by $\mathcal{A}$ but in different clusters by $\mathcal{B}$, 
$N_{01}$ the number of element pairs which are grouped in the same cluster by $\mathcal{B}$ but in different clusters by $\mathcal{A}$, 
and $N_{00}$ the number of element pairs which are grouped in different clusters by both $\mathcal{A}$ and $\mathcal{B}$. 
Intuitively, $N_{11}$ and $N_{00}$ are indicators of the agreement between the two clusterings, while $N_{10}$ and $N_{01}$ reflect the disagreement between the clusterings.

\begin{table}[t!]
 	\begin{center}
    	\begin{tabular}{c|c c c c|c}
	$\mathcal{A} / \mathcal{B}$ & $B_1$ & $B_2$ & $\hdots$ & $B_{K_{\mathcal{B}}}$ & Sums \\ \hline
	$A_1$ & $n_{11}$ & $n_{12}$ & $\hdots$ & $n_{1K_{\mathcal{B}}}$ & $a_1$ \\
	$A_2$ & $n_{21}$ & $n_{22}$ & $\hdots$ & $n_{2K_{\mathcal{B}}}$ & $a_2$ \\
	$\vdots$ & $\vdots$ & $\vdots$ & $\ddots$ & $\vdots$ &$\vdots$ \\
	$A_{K_{\mathcal{A}}}$ & $n_{K_{\mathcal{A}}1}$ & $n_{K_{\mathcal{A}}2}$ & \ldots & $n_{K_{\mathcal{A}}K_{\mathcal{B}}}$ & $a_{K_{\mathcal{A}}}$ \\ \hline
	Sums & $b_{1}$ & $b_{2}$ & $\hdots$ & $b_{K_{\mathcal{B}}}$ & $\sum_{ij}n_{ij} = N$
	\end{tabular}
	\end{center}
	\caption{The contingency table $\mathcal{T}$ for two clusterings $\mathcal{A} = \{A_1, \ldots, A_{K_{\mathcal{A}}}\}$ and $\mathcal{B} = \{B_1,\ldots, B_{K_{\mathcal{B}}}\}$ of $N$ elements, where $n_{ij} = |A_i\cap B_j|$ are the number of elements that are in both cluster $A_i\in\mathcal{A}$ and cluster $B_j\in\mathcal{B}$.}
	\label{tbl:cont}
\end{table}

The aforementioned pair counts are identified from the contingency table $\mathcal{T}$ between two clusterings, shown in \tableref{tbl:cont}, by the following set of equations
\begin{align}
	N_{11} &= \sum_{k,m = 1}^{K_\mathcal{A},K_\mathcal{B}}\binom{n_{km}}{2} = \frac{1}{2}\left(\sum_{k,m = 1}^{K_\mathcal{A},K_\mathcal{B}}n_{km}^2 - N\right) \nonumber \\
	N_{10} &= \sum_{k=1}^{K_\mathcal{A}} \binom{a_k}{2} -N_{11} = \frac{1}{2}\left(\sum_{k=1}^{K_\mathcal{A}} a_k^2 - \sum_{k,m = 1}^{K_\mathcal{A},K_\mathcal{B}}n_{km}^2\right) \\
	N_{01} &= \sum_{m=1}^{K_\mathcal{B}} \binom{b_m}{2} -N_{11} = \frac{1}{2}\left(\sum_{m=1}^{K_\mathcal{B}} b_m^2 - \sum_{k,m = 1}^{K_\mathcal{A},K_\mathcal{B}}n_{km}^2\right) \nonumber \\
	N_{00} &= \binom{N}{2} - N_{11} - N_{10} - N_{01} \nonumber
\end{align}
The Rand index between clusterings $\mathcal{A}$ and $\mathcal{B}$, $\text{RI}(\mathcal{A},\mathcal{B})$ is then given by the function
\begin{align}
	\text{RI}(\mathcal{A},\mathcal{B}) &= \frac{N_{11} + N_{00}}{\binom{N}{2}}  \nonumber \\
	& = \frac{2\sum_{k,m = 1}^{K_\mathcal{A},K_\mathcal{B}}\binom{n_{km}}{2} - \sum_{k= 1}^{K_\mathcal{A}}\binom{a_{k}}{2} - \sum_{m= 1}^{K_\mathcal{B}}\binom{b_{m}}{2} + \binom{N}{2}}{\binom{N}{2}}.
	\label{eq:rand}
\end{align}
It lies between $0$ and $1$, where $1$ indicates the clusterings are identical and $0$ occurs for clusters which do not share a single pair of elements (this only happens when one clustering is the full set of elements and the other clustering groups each element into its own cluster).
As the number of clustered elements increases, the measure becomes dominated by the number of pairs which were classified into different clusters ($N_{00}$), resulting in decreased sensitivity to co-occurring element pairs (\citealp{Fowlkes1983comparehierclusterings}).

Another formulation of the Rand index, used in our later derivations, focuses on a binary representation of the element pairs.
Specifically, consider the vector $U_{\mathcal{A}} =[u_{1}, \ldots, u_{\binom{N}{2}}]$ with binary entries $u_{\alpha}\in\{-1,1\}$ corresponding to all possible element pairs.  Using $\alpha$ to index over all element pairs by $\alpha = \binom{N}{2}-\binom{N - i + 1}{2} + j - i$, for $i<j\leq N$, then $u_{\alpha} = 1$ if elements $v_i$ and $v_j$ are in the same cluster in $\mathcal{A}$ and $u_{\alpha} = -1$ if elements $v_i$ and $v_j$ are in different clusters in $\mathcal{A}$.
There are $Q_{1}^\mathcal{A}$ $1$s in $U_{\mathcal{A}}$ and $Q_{-1}^{\mathcal{A}}$ $-1$s in $U_{\mathcal{A}}$ with
\begin{equation}
	Q_{1}^\mathcal{A} = \sum_{k=1}^{K_{\mathcal{A}}}\binom{a_k}{2}, \hspace{10pt} Q_{-1}^{\mathcal{A}} = \binom{N}{2} -\sum_{k=1}^{K_{\mathcal{A}}}\binom{a_k}{2}.
\end{equation}
The Rand index is found from the vectors $U_{\mathcal{A}}$ and $U_{\mathcal{B}}$, for clusterings $\mathcal{A}$ and $\mathcal{B}$ respectively, as the number of $1$s in their product vector, $U_{\mathcal{A}}\odot U_{\mathcal{B}}$, using element-wise multiplication and normalized by the total size of the vectors, $\binom{N}{2}$.

\subsection{Expected Rand Index, Permutation Model ($M_{\text{perm}}$)}
The expectation of the Rand index with respect to the permutation model follows from drawing the entries in \tableref{tbl:cont} from the generalized hypergeometric distribution.
Utilizing the previous notation with $Q_{1}^\mathcal{A} = \sum_{k=1}^{K_{\mathcal{A}}}\binom{a_k}{2}$, the expectation $\mathbb{E}_{\text{perm}}[RI(\mathcal{A},\mathcal{B})]$ of the Rand index with respect to the permutation model for the cluster size sequences of clusterings $\mathcal{A}$ and $\mathcal{B}$ is given by
\begin{equation}
	\mathbb{E}_{\text{perm}}[\text{RI}(\mathcal{A},\mathcal{B})] =  \frac{2Q_{1}^\mathcal{A}Q_{1}^\mathcal{B} -  \binom{N}{2}\Big(Q_{1}^\mathcal{A} + Q_{1}^\mathcal{B}\Big) + \binom{N}{2}^2}{\binom{N}{2}^2}
	\label{eq:hyperexprand}
\end{equation}
(see \citealp{Fowlkes1983comparehierclusterings}, \citealp{Hubert1985adjustedrand}, or \citealp{Albatineh2011correctingpairwise} for the full derivation).

The commonly used adjusted Rand index (ARI) of \cite{Hubert1985adjustedrand} uses $M_{\text{perm}}$ to calculate the expectation of the Rand index, $\mathbb{E}_{\text{perm}}[\text{RI}(\mathcal{A},\mathcal{B})]$, as found in \equationref{eq:hyperexprand}.  This expectation is then used in \equationref{eq:adjchance}, along with the fact that the maximum value of the Rand index is $\max_{\text{perm}}[\text{RI}] = 1.0$, to give
\begin{equation}
	\text{ARI}_{\text{perm}}(\mathcal{A},\mathcal{B}) = \frac{\binom{N}{2} \sum_{k,m = 1}^{K_{\mathcal{A}}K_{\mathcal{B}}}\binom{n_{km}}{2}-\sum_{k=1}^{K_{\mathcal{A}}}\binom{a_k}{2}\sum_{m=1}^{K_{\mathcal{B}}}\binom{b_m}{2}}{\frac{1}{2}\binom{N}{2}\left[\sum_{k=1}^{K_{\mathcal{A}}}\binom{a_k}{2} + \sum_{m=1}^{K_{\mathcal{B}}}\binom{b_m}{2}\right] - \sum_{k=1}^{K_{\mathcal{A}}}\binom{a_k}{2}\sum_{m=1}^{K_{\mathcal{B}}}\binom{b_m}{2}}.
\end{equation}

\subsection{Expected Rand Index, Fixed Number of Clusters}
We follow \cite{Dubien1981extendedrandmodels} to calculate the Rand index between two clusterings under the assumptions that both clusterings were independently and uniformly drawn from the ensemble of clusterings with a fixed number of clusters ($M_{\text{num}}$).
Recall that the Rand index between two clusterings $\mathcal{A}$ and $\mathcal{B}$ is given by the number of $1$s in the element-wise product of the binary representations vectors $U_{\mathcal{A}}$ and $U_{\mathcal{B}}$.
The expected Rand index under any random model is then the expected number of $1$s in this product vector, normalized by the total size of the vector
\begin{align}
	\label{eq:anymeanrand}
	\mathbb{E}[RI(\mathcal{A},\mathcal{B})] &= \mathbb{E}\left[\frac{1}{\binom{N}{2}}\sum_{\alpha = 1}^{\binom{N}{2}}\bm{1}_{u_\alpha^{\mathcal{A}} \cdot u_\alpha^{\mathcal{B}} = 1}\right] \\
	&=\frac{1}{\binom{N}{2}}\sum_{\alpha = 1}^{\binom{N}{2}} \mathbb{E}\left[\bm{1}_{u_\alpha^{\mathcal{A}} \cdot u_\alpha^{\mathcal{B}} = 1}\right] \nonumber \\
	&=\frac{1}{\binom{N}{2}}\sum_{\alpha = 1}^{\binom{N}{2}} P(u_\alpha^{\mathcal{A}}\cdot u_{\alpha}^{\mathcal{B}} = 1) \nonumber 
\end{align}
The product $u_\alpha^{\mathcal{A}}\cdot u_{\alpha}^{\mathcal{B}} $ equals $1$ when either $u_\alpha^{\mathcal{A}} = 1$ and $u_{\alpha}^{\mathcal{B}} = 1$, or $u_\alpha^{\mathcal{A}} = -1$ and $u_{\alpha}^{\mathcal{B}} = -1$.
Since we assumed both clusterings were independent, this gives
\begin{equation}
	P(u_\alpha^{\mathcal{A}}\cdot u_{\alpha}^{\mathcal{B}} = 1) = P(u_\alpha^{\mathcal{A}} = 1)P(u_{\alpha}^{\mathcal{B}} = 1) + P(u_\alpha^{\mathcal{A}} = -1)P(u_{\alpha}^{\mathcal{B}} = -1)
	\label{eq:randprob}
\end{equation}
where $P(u_\alpha^{\mathcal{A}} = 1)$ is the probability that the two elements $v_i$ and $v_j$ are in the same cluster in clustering $\mathcal{A}$, where the element pair is indexed by $\alpha = \binom{N}{2}-\frac{(N-i)(N-i+1)}{2} + j - i$ with $i<j\leq N$.
Likewise, $P(u_\alpha^{\mathcal{A}} = -1)$ is the probability that the two elements $v_i$ and $v_j$ are in different clusters.

Under the assumption of $M_{\text{num}}$, there is a uniform probability of selecting a clustering from the $\stirling{N}{K_{\mathcal{A}}}$ clusterings of $N$ elements into $K_{\mathcal{A}}$ clusters; we define, $P_{\text{num}}(u_\alpha^{\mathcal{A}} = 1)$ as the proportion of these clusterings with elements $v_i$ and $v_j$ in the same cluster.
To find this proportion, notice that we can ensure $v_i$ is in the same cluster as $v_j$ by first partitioning all elements besides $v_i$ into $K_{\mathcal{A}}$ clusters; then, we can add $v_i$ to the same cluster as $v_j$.
Since there are $\stirling{N-1}{K_{\mathcal{A}}}$ such clusterings without element $v_i$, this gives 
\begin{align}
    \label{eq:pnumrand}
    P_{\text{num}}(u_{\alpha}^{\mathcal{A}} = 1) &= \frac{\stirling{N-1}{K_{\mathcal{A}}}}{\stirling{N}{K_{\mathcal{A}}}} \\
    P_{\text{num}}(u_{\alpha}^{\mathcal{A}} = -1) &= 1 - \frac{\stirling{N-1}{K_{\mathcal{A}}}}{\stirling{N}{K_{\mathcal{A}}}}.
\end{align}

Finally, the expected Rand index between two clusterings $\mathcal{A}$ and $\mathcal{B}$ with $K_{\mathcal{A}}$ and $K_{\mathcal{B}}$ clusters assuming $M_{\text{num}}$ is given by
\begin{align}
	\mathbb{E}_{\text{num}}[\text{RI}(\mathcal{A},\mathcal{B})] &=  \frac{\stirling{N-1}{K_\mathcal{A}}}{\stirling{N}{K_A}} \frac{\stirling{N-1}{K_\mathcal{B}}}{\stirling{N}{K_\mathcal{B}} } \nonumber \\
	&+ \left(1 - \frac{\stirling{N-1}{K_\mathcal{A}} }{\stirling{N}{K_\mathcal{A}} }\right)\left(1 - \frac{\stirling{N-1}{K_\mathcal{B}} }{\stirling{N}{K_\mathcal{B}} }\right).
	\label{eq:numrandmean}
\end{align}

When $N$ is large, we can approximate the Stirling numbers of the second kind for a fixed $K$ by $\stirling{N}{K} \approx \frac{K^N}{K!}$.
This can be inserted into equation \eqref{eq:numrandmean} to give the following approximation for the mean of the Rand index assuming $M_{\text{num}}$
\begin{equation}
	\mathbb{E}_{\text{num}}[\text{RI}(\mathcal{A},\mathcal{B})] \approx \frac{1}{K_{\mathcal{A}}K_{\mathcal{B}}}+\left(1-\frac{1}{K_{\mathcal{A}}}\right)\left(1-\frac{1}{K_{\mathcal{B}}} \right).
\end{equation}
Interestingly, this suggests that the Rand index goes to $1$ at a rate inversely related to the smaller number of clusters $\mathcal{O}(\max\{K_{\mathcal{A}}^{-1}, K_{\mathcal{B}}^{-1}\})$.

\subsection{Expected Rand Index, All Clusterings $M_{\text{all}}$}
The average of the Rand index between two clusterings under the assumption that the clusterings were drawn with uniform probability from the set of all clusterings directly follows from the random model with a fixed number of clusters previously discussed.
Namely, because Bell numbers are related to Stirling numbers of the second kind by $B_N = \sum_{k=1}^N\stirling{N}{k}$, a similar reasoning as followed for equation \eqref{eq:pnumrand} gives
\begin{align}
	P_{\text{all}}(u_{\alpha} = 1) &= \sum_{k=1}^N\frac{\stirling{N}{k}}{B_N}P_{\text{num}}(u_{\alpha}^{k} = 1) \\
	&= \frac{1}{B_N}\sum_{k=1}^N\stirling{N}{k} \frac{\stirling{N-1}{k}}{\stirling{N}{k}} \nonumber \\
	&= \frac{B_{N-1}}{B_N}.
	\label{eq:randallterm}
\end{align}
Using this probability for the expectation in equation \eqref{eq:anymeanrand} gives the expected Rand index under the assumption that both clusterings were uniformly drawn from the set of all clusterings of $N$ elements
\begin{align}
	\mathbb{E}_{\text{all}}[\text{RI}(\mathcal{A},\mathcal{B})] &= \left(\frac{B_{N-1}}{B_N}\right) ^2 + \left(1 - \frac{B_{N-1}}{B_N}\right)^2.
	\label{eq:exprandall}
\end{align}

When $N$ is large, we can approximate the ratio of successive Bell numbers by $\frac{B_{N+1}}{B_N} \approx \frac{N}{\log N}$.  Using this approximation in equation \eqref{eq:exprandall} gives the following approximation for the mean of the Rand index in $M_{\text{all}}$
\begin{equation}
	\mathbb{E}_{\text{all}}[\text{RI}(\mathcal{A},\mathcal{B})] \approx \left(\frac{\log N}{N}\right)^2+\left(1-\frac{\log N}{N}\right)^2.
\end{equation}
Interestingly, this suggests that the expected Rand index between two random clusterings goes to $1$ at a rate $\mathcal{O}\left(\frac{\log(N)}{N}\right)$, inversely proportional to the number of elements.

\subsection{One-Sided Rand}
Consider a reference clustering $\mathcal{G}$ that has the cluster size sequence $[g_1, \ldots, g_{K_{\mathcal{G}}}]$.  The binary pair vector representation of $\mathcal{G}$ has $Q_{1}^{\mathcal{G}}$ $1$s and $Q_{-1}^{\mathcal{G}} = \binom{N}{2} - Q_{1}^{\mathcal{G}}$, $-1$s.
The one-sided expectation of the Rand index under the assumption that clustering $\mathcal{A}$ was randomly drawn from either the $M_{\text{num}}$ or $M_{\text{all}}$ random models follows from treating the two  clusterings independently as in equation \eqref{eq:randprob}.
Since the cluster sequence for the reference clustering is fixed, the probability that a random entry in the binary pair vector is $1$ is given by the fraction of $1$s in the vector
\begin{align}
	P_{\text{num}}(u_{\alpha}^{\mathcal{G}} = 1)  &= \frac{1}{\binom{N}{2}}Q_{1}^{\mathcal{G}} \\
		&= \frac{1}{\binom{N}{2}}\sum_{i=1}^{K_{\mathcal{G}}}\binom{g_i}{2}. \nonumber
\end{align}
The one-sided expectation of the Rand index under the assumption that clustering $\mathcal{A}$ was randomly drawn from the set of all clusterings with a fixed number of clusters $M_{\text{num}}^1$ is
\begin{align}
	\mathbb{E}_{\text{num}}^1[\text{RI}(\mathcal{A},\mathcal{G})] &= \left(\frac{\stirling{N-1}{K_{\mathcal{A}}} }{\stirling{N}{K_{\mathcal{A}}}} \frac{Q_{1}^\mathcal{G}}{\binom{N}{2}}\right) + \left(1 - \frac{\stirling{N-1}{K_{\mathcal{A}}}}{\stirling{N}{K_{\mathcal{A}}}}\right) \left(1 - \frac{Q_{1}^\mathcal{G}}{\binom{N}{2}}\right).
\end{align}
The one-sided expectation of the Rand index with the assumption that the random clustering $\mathcal{A}$ is drawn from the ensemble of all partitions $M_{\text{all}}^1$ is
\begin{align}
	\mathbb{E}_{\text{all}}^1[\text{RI}(\mathcal{A},\mathcal{G})] &= \frac{B_{N-1}}{B_{N}} \frac{Q_{1}^{\mathcal{G}}}{\binom{N}{2}} + \left(1 - \frac{B_{N-1}}{B_N}\right)\left(1 - \frac{Q_{1}^\mathcal{G}}{\binom{N}{2}}\right).
\end{align}

\section{Mutual Information}
\label{sec:mi}
Another prominent family of clustering similarity measures is based on the Shannon information between probabilistic representations of each clustering.
These probability distributions are also calculated from the contingency table $\mathcal{T}$, \tableref{tbl:cont}.
The partition entropy $H$ of a clustering $\mathcal{A}$ is given by
\begin{equation}
		H(\mathcal{A}) = - \sum_{k=1}^{K_\mathcal{A}}\frac{a_k}{N}\log\frac{a_k}{N}.
\end{equation}
Using this entropy, the mutual information $\text{MI}(\mathcal{A},\mathcal{B})$ between two clusterings $\mathcal{A}$ and $\mathcal{B}$ is given by
\begin{align}
	\text{MI}(\mathcal{A}, \mathcal{B}) &= H(\mathcal{A}) + H(\mathcal{B}) - H(\mathcal{A}, \mathcal{B}) \nonumber \\
			&= \sum_{k,m = 1}^{K_{\mathcal{A}}, K_{\mathcal{B}}}\frac{n_{km}}{N}\log\frac{n_{km} N}{a_k b_m}.
\end{align}
The mutual information can be interpreted as an inverse measure of independence between the clusterings, or a measure of the amount of information each clustering has about the other.
As it can vary in the range $[0,\min\{H(\mathcal{A}),H(\mathcal{B})\}]$, to facilitate comparisons, it is desirable to normalize it to the range $[0,1]$.  There are at least six proposals in the literature for this upper bound, each with different advantages and drawbacks
\begin{align}
	\label{eq:mimax}
	\min\{H(\mathcal{A}),H(\mathcal{B})\} \leq \sqrt{H(\mathcal{A})H(\mathcal{B})} \leq \frac{H(\mathcal{A}) + H(\mathcal{B})}{2}  \\ 
	 \leq \max\{H(\mathcal{A}),H(\mathcal{B})\} \leq \max\{\log K_{\mathcal{A}}, \log K_{\mathcal{B}}\} \leq \log N. \nonumber
\end{align}
The resulting measures are all known as normalized mutual information (NMI).
This measure has been said to exhibit more desirable properties than the Rand index; for example, it is dependent on the relative proportions of the cluster sizes in each clustering rather than the number of elements.  However, due to its dependence on the number of clusters in each clustering, it is known to favor comparisons between clusterings with more clusters regardless of any other shared clustering features (\citealp{White1994nmiproblems, Vinh2010correctionnmi, Amelio2015nmifair}).

\subsection{Expected Mutual Information, Permutation Model ($M_{\text{perm}}$)}
The mutual information between two clusterings has also previously been studied under the assumption that both clusterings were randomly generated from the permutation model (\citealp{Vinh2009correctingnmi, Romano2014correctingnmi, Vinh2010correctionnmi}).
Expanding the definition of the mutual information gives
\begin{align}
	\mathbb{E}_{\text{perm}}[\text{MI}(\mathcal{A},\mathcal{B})] &= \mathbb{E}_{\text{perm}}[H(\mathcal{A})] + \mathbb{E}_{\text{perm}}[H(\mathcal{B})] - \mathbb{E}_{\text{perm}}[H(\mathcal{A}, \mathcal{B})] \\
	&= H(\mathcal{A}) + H(\mathcal{B}) - \mathbb{E}_{\text{perm}}[H(\mathcal{A}, \mathcal{B})] \nonumber
	\label{eq:expectmiperm}
\end{align}
where the second line follows from the fact that all cluster sizes (and hence the entropy) are the same for every clustering in $M_{\text{perm}}$.

The expectation of the joint entropy with respect to $M_{\text{perm}}$ for the cluster size distributions of clusterings $\mathcal{A}$ and $\mathcal{B}$ is the average over all possible contingency tables $\mathcal{T}$ with entries $n$
\begin{align}
	\mathbb{E}_{\text{perm}}[H(\mathcal{A},\mathcal{B})] = -\sum_{\mathcal{T}}p(\mathcal{T}|\mathcal{A},\mathcal{B})\sum_{k=1}^{K_{\mathcal{A}}}\sum_{m=1}^{K_{\mathcal{B}}}\frac{n}{N}\log\left(\frac{n}{N}\right).
\end{align}
Rearranging the summations, and recalling that the entries of the contingency tables are hyper-geometrically distributed such that the probability of each entry 
\begin{equation}
p(n) = \frac{\binom{b_m}{n}\binom{N-b_{m}}{a_k - n}}{\binom{N}{a_k}}
\end{equation}
is only dependent on the row sum $a_k$ and column sum $b_m$, gives
\begin{align}
	\mathbb{E}_{\text{perm}}[H(\mathcal{A},\mathcal{B})] = -\sum_{k=1}^{K_{\mathcal{A}}}\sum_{m=1}^{K_{\mathcal{B}}}\sum_{n}\frac{n}{N}\log\left(\frac{n}{N}\right)\frac{\binom{b_m}{n}\binom{N-b_{m}}{a_k - n}}{\binom{N}{a_k}}.
\end{align}
According to the hyper-geometric distribution, the summation over table entries $n_{km}$ occurs between the lower bound: $\max\{0, a_k + b_m - N\}$ and the upper bound: $\min\{a_k,b_m\}$.
Combining this expression with the individual entropies $H(\mathcal{A})$ and $H(\mathcal{B})$ gives (\citealp{Vinh2009correctingnmi})
\begin{align}
	\mathbb{E}_{\text{perm}}[\text{MI}(\mathcal{A},\mathcal{B})] = \sum_{k=1}^{K_{\mathcal{A}}}\sum_{m=1}^{K_{\mathcal{B}}}\sum_{n}\frac{n}{N}\log\left(\frac{Nn}{a_kb_m}\right)\frac{\binom{b_m}{n}\binom{N-b_{m}}{a_k - n}}{\binom{N}{a_k}}.
	\label{eq:expectnmipermexpand}
\end{align}
As shown in \cite{Romano2014correctingnmi}, the computational complexity of calculating the expected mutual information assuming the permutation model is of order $\mathcal{O}(\max\{K_{\mathcal{A}}N,K_{\mathcal{B}}N\})$.

The adjusted mutual information (AMI) of \cite{Vinh2009correctingnmi} uses $M_{\text{perm}}$ to correct the MI for chance according to equation \eqref{eq:adjchance} and selecting an upper bound $\max[\text{MI}]$ from equation \eqref{eq:mimax} to give
\begin{equation}
	\text{AMI}(\mathcal{A},\mathcal{B}) = \frac{\text{MI}(\mathcal{A},\mathcal{B}) - \mathbb{E}_{\text{perm}}[\text{MI}(\mathcal{A},\mathcal{B})]}{\max_{\text{perm}}[\text{MI}] - \mathbb{E}_{\text{perm}}[\text{MI}(\mathcal{A},\mathcal{B})]}.
\end{equation}

\subsection{Expected Mutual Information, Fixed Number of Clusters ($M_{\text{num}}$)}
Next, we consider the Mutual Information between two clusterings under the assumptions that both clusterings were independently and uniformly drawn from the ensemble of clusterings with a fixed number of clusters ($M_{\text{num}}$).
In this case, the expected mutual information is dependent on both the average partition entropy and the joint partition entropy
\begin{equation}
	\mathbb{E}_{\text{num}}[\text{MI}(\mathcal{A},\mathcal{B})] = \mathbb{E}_{\text{num}}[H(\mathcal{A})] + \mathbb{E}_{\text{num}}[H(\mathcal{B})] - \mathbb{E}_{\text{num}}[H(\mathcal{A}, \mathcal{B})].
	\label{eq:expectminum}
\end{equation}
This expectation can be found by considering the average partition entropy and joint partition entropy separately.
Recall that in the permutation model $\mathbb{E}_{\text{perm}}[H(\mathcal{A})] = H(\mathcal{A})$ since the cluster sizes remain unchanged; however, the same does not hold in $M_{\text{num}}$.
Denoting a random clustering with $K_{\mathcal{A}}$ clusters as $\pi_{K_{\mathcal{A}}}$, and using the notion $\sum_{\sigma_i\in \pi_{K_{\mathcal{A}}}}$ to indicate the summation over all clusters in the clustering $\pi_{K_{\mathcal{A}}}$, where the cardinality of the cluster is $|\sigma_i| = a$, then the expected partition entropy of a random clustering in $M_{\text{num}}$ is
\begin{align}
	\mathbb{E}_{\text{num}}[H(\mathcal{A})] &= -\sum_{\pi_{K_\mathcal{A}}}p_{\text{num}}(\pi_{K_\mathcal{A}})\sum_{\sigma_i\in\pi_{K_\mathcal{A}}}\frac{a}{N}\log\left(\frac{a}{N}\right).
\end{align}
Note that this expression only depends on the size $a = |\sigma_i|$ for $\sigma_i\in\pi_{K_\mathcal{A}}$ of the clusters in the clustering.  This means the expected entropy of a random clustering can be rewritten in terms of the expected contribution to the entropy from a random cluster of size $a$.
A counting argument gives the number of clusters of size $a$ which appear in all of the clusterings in the random ensemble.
First, choose $a$ of the $N$ elements to form the cluster.
Each clustering in $M_{\text{num}}$ must have $K_{\mathcal{A}}$ clusters, so the remaining $N-a$ elements have to be arranged into $K_{\mathcal{A}} -1$ other clusters.
There are $\stirling{N-a}{K_{\mathcal{A}} -1}$ ways to partition these remaining elements.
This gives $\binom{N}{a}\stirling{N-a}{K_{\mathcal{A}} -1}$ clusters of size $a$ (\citealp{Chern2014expectedsetpartitions}).
The expected number of clusters $n_{\text{num}}(a)$ in a random clustering drawn from $M_{\text{num}}$ is then $\binom{N}{a}\frac{\stirling{N-a}{K_{\mathcal{A}} -1}}{\stirling{N}{K_{\mathcal{A}}}}$.
Therefore, the expected clustering entropy in $M_{\text{num}}$ is:
\begin{align}
	\mathbb{E}_{\text{num}}[H(\mathcal{A})] &= -\sum_{a = 1}^{N - (K_{\mathcal{A}} -1)}  \binom{N}{a}\frac{\stirling{N-a}{K_{\mathcal{A}} -1}}{\stirling{N}{K_{\mathcal{A}}}} \frac{a}{N}\log\left(\frac{a}{N}\right).
\end{align}
where the summation is over all possible cluster sizes $[1, N - (K_{\mathcal{A}} -1)]$ encountered when partitioning $N$ elements into $K_{\mathcal{A}}$ clusters.

Similarly, the expected joint entropy of two random clusterings drawn independently from $M_{\text{num}}$ is given by the expected number of clusters of size $a$ from a clustering with $K_{\mathcal{A}}$ clusters, the expected number of clusters of size $b$ from a clustering with $K_{\mathcal{B}}$ clusters, and then considering the probability of overlap $p(n_{km}) = \frac{\binom{b}{n_{km}}\binom{N-b}{a - n_{km}}}{\binom{N}{a}}$ from the resulting random contingency table
\begin{align}
	\mathbb{E}_{\text{num}}[H(\mathcal{A},\mathcal{B})] &= -\sum_{\pi_{K_\mathcal{A}}}p_{\text{num}}(\pi_{K_\mathcal{A}})\sum_{\pi_{K_\mathcal{B}}}p_{\text{num}}(\pi_{K_\mathcal{B}} ) \nonumber \\ 
	& \times\sum_{k=1}^{K_{\mathcal{A}}}\sum_{m=1}^{K_{\mathcal{B}}}\sum_{n}\frac{n}{N}\log\left(\frac{Nn}{a_kb_m}\right)\frac{\binom{b_m}{n}\binom{N-b_{m}}{a_k - n}}{\binom{N}{a_k}} \nonumber \\
	&=- \sum_{a = 1}^{N - (K_{\mathcal{A}}- 1)}  \sum_{b = 1}^{N - (K_{\mathcal{B}}-1)} \sum_{n} \left[\binom{N}{a}\frac{\stirling{N-a}{K_{\mathcal{A}} -1}}{\stirling{N}{K_{\mathcal{A}}}} \right. \\ 
	& \times \left. \binom{N}{b}\frac{\stirling{N-b}{K_{\mathcal{B}} -1}}{\stirling{N}{K_{\mathcal{B}}}}\frac{n}{N} \log\left(\frac{n}{N}\right)\frac{\binom{b}{n}\binom{N-b}{a - n}}{\binom{N}{a}} \right]. \nonumber
	\label{eq:expectnmi}
\end{align}
Note that, when using equation \eqref{eq:adjchance} to adjust the mutual information for chance under the assumption of $M_{\text{num}}$, the maximum value for the measure over the entire ensemble of random clusterings has to be used.
When considering clusterings with a fixed number of clusters, we know that $H(\mathcal{A})\leq \log K_{\mathcal{A}}$.
This means that the choices for $\max_{\text{num}}[\text{MI}(\mathcal{A},\mathcal{B})]$ are
\begin{equation}
 \min\{\log K_{\mathcal{A}}, \log K_{\mathcal{B}}\} \leq \sqrt{\log K_{\mathcal{A}} \log K_{\mathcal{B}}}  \leq \frac{1}{2}\log K_{\mathcal{A}}K_{\mathcal{B}}\leq \max\{\log K_{\mathcal{A}}, \log K_{\mathcal{B}}\}.
\end{equation}

As is apparent from the summations in equation (32), the computational complexity of exactly calculating the expected mutual information assuming $M_{\text{num}}$ is of order $\mathcal{O}(N^3)$.

\subsection{Expected Mutual Information, All Clusterings $M_{\text{all}}$}
The expected Mutual Information between two clusterings under the assumption that both clusterings were independently and uniformly drawn from the set of all possible clusterings of $N$ elements, $M_{\text{all}}$, has a similar derivation as the previous case of $M_{\text{num}}$.
Both the expectations for the entropy of a single clustering and the joint entropy of the two clusterings need to be considered separately and can be rewritten in terms of the contributions from individual clusters of a given size.
In $M_{\text{all}}$, the number of clusters of size $a$ is again found by choosing $a$ of the $N$ elements for the cluster and then partitioning the remaining $N-a$ elements; there are now $B_{N-a}$ possible ways to cluster the remaining elements (\citealp{Chern2014expectedsetpartitions}).
This gives
\begin{align}
	\mathbb{E}_{\text{all}}[H(\mathcal{A})] &= -\sum_{a = 1}^{N}   \binom{N}{a}\frac{B_{N-a}}{B_N} \frac{a}{N}\log\left(\frac{a}{N}\right).
\end{align}
The expected joint entropy for two clusterings is then
\begin{align}
	\mathbb{E}_{\text{all}}[H(\mathcal{A},\mathcal{B})] &= -\sum_{\pi_a}p(\pi_a)\sum_{\pi_b}p(\pi_b)\sum_{i=1}^{N}\sum_{j=1}^{N}\sum_{n}\frac{n}{N}\log\left(\frac{Nn}{a_ib_j}\right)\frac{\binom{b_j}{n}\binom{N-b_{j}}{a_i - n}}{\binom{N}{a_i}} \nonumber \\
	&= -\sum_{a = 1}^{N}  \binom{N}{a}\frac{B_{N-a}}{B_{N}} \sum_{b = 1}^{N}   \binom{N}{b}\frac{B_{N-b}}{B_{N}} \sum_{n}\frac{n}{N}\log\left(\frac{n}{N}\right)\frac{\binom{b}{n}\binom{N-b}{a - n}}{\binom{N}{a}}  \nonumber \\
	&= -2\sum_{a = 1}^{N} \sum_{b = 1}^{a-1}  \binom{N}{a}\frac{B_{N-a}}{B_{N}}  \binom{N}{b}\frac{B_{N-b}}{B_{N}}\sum_{n}\frac{n}{N}\log\left(\frac{n}{N}\right)\frac{\binom{b}{n}\binom{N-b}{a - n}}{\binom{N}{a}} \\
	&- \sum_{a = 1}^{N} \left( \binom{N}{a}\frac{B_{N-a}}{B_{N}} \right)^2\sum_{n}\frac{n_{ij}}{N}\log\left(\frac{n}{N}\right)\frac{\binom{a}{n}\binom{N-a}{a - n}}{\binom{N}{a}}, \nonumber
\end{align}
with the last simplification resulting from the symmetry of the hyper-geometric term with respect to $a$ and $b$.

As in the previous case, the maximum bound of the measure must be consider over the entire ensemble of clusterings.
Again, we consider the bound $H(\mathcal{A}) \leq \log N$.
This reduces to only one choice for $\max_{\text{all}}[\text{MI}(\mathcal{A},\mathcal{B})] = \log N$.

\subsection{One-Sided Mutual Information}
As was the case for the one-sided expected Rand index, the one-sided expectation of mutual information follows from the fact that the cluster sequence for the reference clustering is fixed.
This results in the following one-sided expected joint entropy when the random clustering $\mathcal{A}$ is drawn from the $M_{\text{num}}$ model
\begin{align}
	\mathbb{E}_{\text{num}}^1[H(\mathcal{A},\mathcal{G})] &=  -\sum_{a = 1}^{N}  \binom{N}{a}\frac{\stirling{N-a}{K_{\mathcal{A}}-1}}{ \stirling{N}{K_{\mathcal{A}}}} \sum_{b = 1}^{K_{\mathcal{G}}}\sum_{n}\frac{n}{N}\log\left(\frac{n}{N}\right)\frac{\binom{g_{b}}{n}\binom{N-g_{b}}{a - n}}{\binom{N}{a}}.
\end{align}
The corresponding one-sided expected MI assuming $M_{\text{num}}^1$ is
\begin{align}
	\label{eq:expect1mi}
	\mathbb{E}_{\text{num}}^1[\text{MI}(\mathcal{A},\mathcal{G})] &= -\sum_{a=1}^{K_{\mathcal{A}}}\binom{N}{a}\frac{\stirling{N-a}{K_{\mathcal{A}}-1}}{\stirling{N}{K_{\mathcal{A}}}}\frac{a}{N}\log\left(\frac{a}{N}\right) - \sum_{b=1}^{K_{\mathcal{G}}}\frac{g_b}{N}\log\left(\frac{g_b}{N}\right) \\
	&+ \sum_{a = 1}^{N}  \binom{N}{a}\frac{\stirling{N-a}{K_{\mathcal{A}}-1}}{ \stirling{N}{K_{\mathcal{A}}}} \sum_{b = 1}^{K_{\mathcal{G}}}\sum_{n} \frac{n}{N}\log\left(\frac{n}{N}\right)\frac{\binom{g_{b}}{n}\binom{N-g_{b}}{a - n}}{\binom{N}{a}}. \nonumber
\end{align}
The one-sided expected joint entropy when the random clustering $\mathcal{A}$ is drawn from the $M_{\text{all}}^1$ model is
\begin{align}
	\mathbb{E}_{\text{all}}^1[H(\mathcal{A},\mathcal{G})] &=   -\sum_{a = 1}^{N}  \binom{N}{a}\frac{B_{N-a}}{B_{N}} \sum_{b = 1}^{K_{\mathcal{G}}}\sum_{n}\frac{n}{N}\log\left(\frac{n}{N}\right)\frac{\binom{g_{b}}{n}\binom{N-g_{b}}{a - n}}{\binom{N}{a}},
	\label{eq:1alljointentropy}
\end{align}
and the one-sided expectation of the MI when the random clustering $\mathcal{A}$ is drawn from the $M_{\text{all}}^1$ model is
\begin{align}
	\mathbb{E}_{\text{all}}^1[\text{MI}(\mathcal{A},\mathcal{G})] &=- \sum_{a=1}^{K_{\mathcal{A}}}\binom{N}{a}\frac{B_{N-a}}{B_{N}}\frac{a}{N}\log\left(\frac{a}{N}\right) - \sum_{b=1}^{K_{\mathcal{G}}}\frac{g_b}{N}\log\left(\frac{g_b}{N}\right) \\
	&+ \sum_{a = 1}^{N}  \binom{N}{a}\frac{B_{N-a}}{B_{N}} \sum_{b = 1}^{K_{\mathcal{G}}}\sum_{n}\frac{n}{N}\log\left(\frac{n}{N}\right)\frac{\binom{g_{b}}{n}\binom{N-g_{b}}{a - n}}{\binom{N}{a}}. \nonumber
	\label{eq:1alljmi}
\end{align}
Again, the maximum bound must be chosen with respect to the measure maximum over the clusterings present in the random model.

\section{Results}
\label{sec:results}
The choice of random model for clusterings and the choice of one-sided comparisons can significantly affect results of clustering comparisons.
We first illustrate that the ranking of similar clustering pairs (or, equivalently, finding the most similar clustering pair) depends on the choices of random models in a hypothetical example (\secref{sec:hypoexample}) and K-means clustering of a handwritten digits data set (\secref{sec:kmeanexample}).
One of the primary reasons such strong discrepancies occur is that the cluster size sequences are fixed within samples from $M_{\text{perm}}$.
This means that adjusted comparisons using $M_{\text{perm}}$ are unable to differentiate random clusterings with drastically different cluster size sequences, as we illustrate through our third example in \secref{sec:paexample}.
Second, we demonstrate that the interpretation of adjusted clustering similarity measures with respect to a random baseline also depends on the random model through an evaluation of hierarchical clustering applied to gene expression data in \secref{sec:geneexample}.
Crucially, all of these examples illustrate that conclusions based on corrected similarity measures can change depending on the random model for clusterings.

\subsection{Clustering Similarity Ranking}
\label{sec:hypoexample}
\begin{figure}[t]
	\centering
	\includegraphics[width = \columnwidth]{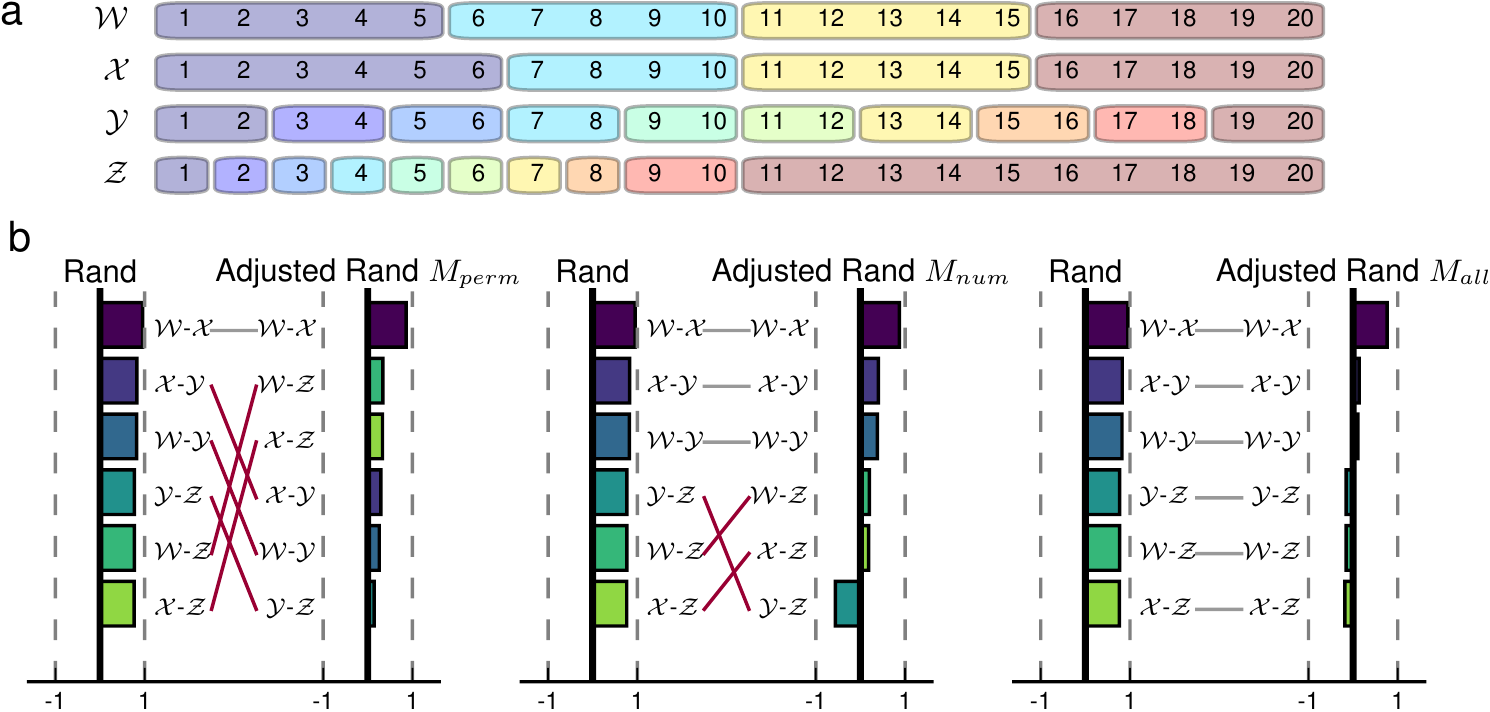}
	\caption{The choice of random model for the Rand index has a significant impact on the rankings of clustering similarity.  \textbf{a}, Four clusterings of $N=20$ elements; $\mathcal{W}$ and $\mathcal{X}$ each contain four clusters and differ by the assignment of one element ($6$), $\mathcal{Y}$ and $\mathcal{Z}$ each contain ten clusters.  \textbf{b}, Rankings for the similarity of clustering pairs using the Rand index, the Adjusted Rand index assuming $M_{\text{perm}}$, the Adjusted Rand index assuming $M_{\text{num}}$, and the Adjusted Rand index assuming $M_{\text{all}}$. Rankings which change as a function of random model are highlighted in dark red.}
	\label{fig:randranking}
\end{figure}

\emph{Our first example demonstrates how rankings assigned by the similarity score can change depending on the assumed random model.}
Consider the four hypothetical clusterings of $20$ elements presented in \figureref{fig:randranking}a.
Clustering $\mathcal{W}$ contains four equally sized clusters; clustering $\mathcal{X}$ is generated by shifting the membership of one element from $\mathcal{W}$; clustering $\mathcal{Y}$ groups the elements into $10$ equally sized clusters; and clustering $\mathcal{Z}$ groups the elements into $10$ heterogeneous clusters.
The similarity (from the most similar at the top to the least similar at the bottom) of all 6 clustering pairs is ranked using the Rand index and each of its three adjusted variants in \figureref{fig:randranking}b.  
Note that the adjusted Rand index can be negative.
The unadjusted Rand index ranking serves as a reference to illustrate how the random models change rankings.
%

As one would expect, all four Rand measures identify clusterings $\mathcal{W}$ and $\mathcal{X}$ as the most similar (\figureref{fig:randranking}b).
However, the ranking of the other five comparisons varies widely as a result of the underlying random models.
These changes can be understood by tracking comparisons to clustering $\mathcal{Z}$.
The low Rand index for the three comparisons with clustering $\mathcal{Z}$ ($\approx0.76$) reflects the fact that clustering $\mathcal{Z}$ has a drastically different number of clusters or cluster size sequence from the other three clusterings.
The permutation model retains these differences in all random clusterings; the resulting adjusted index thus treats comparisons between clustering $\mathcal{Z}$ and either $\mathcal{W}$ or $\mathcal{X}$ more favorably than those to clustering $\mathcal{Y}$.
On the other hand, the cluster size sequence for clustering $\mathcal{Z}$ is relatively rare in both $M_{\text{num}}$ and $M_{\text{all}}$.
Since clusterings $\mathcal{Y}$ and $\mathcal{Z}$ have the same number of clusters, the differences in their adjusted scores using $M_{\text{num}}$ are a consequence of their cluster size sequences.
Finally, all four clusterings are over $20$ elements---the only factor that specifies the expected Rand index assuming $M_{\text{all}}$---so they are all adjusted by the same amount when $M_{\text{all}}$ is used.
Note that in our example, clustering $\mathcal{Z}$ has a negative adjusted Rand score using the $M_{\text{all}}$ when compared with all three other clusterings, thus it is less similar to the other three clusterings than one would have expected from comparing two completely random clusterings.

This example illustrates an important property of the $M_{\text{all}}$ model.
Namely, the ranking provided by the Rand index remains unchanged whenever the $M_{\text{all}}$ model is used for adjustment because all clusterings have the same number of elements.
However, the corrected baseline now provides a strong interpretation for negative scores: Two randomly selected clusterings are expected to be more similar.
This is an important consideration for the evaluation of clustering methods; if the derived clustering is no more similar than would be expected when comparing completely random clusterings, the solution is likely not a meaningful representation of the data.

\afterpage{
\begin{figure}[ht]
	\centering
	\includegraphics[width = \columnwidth]{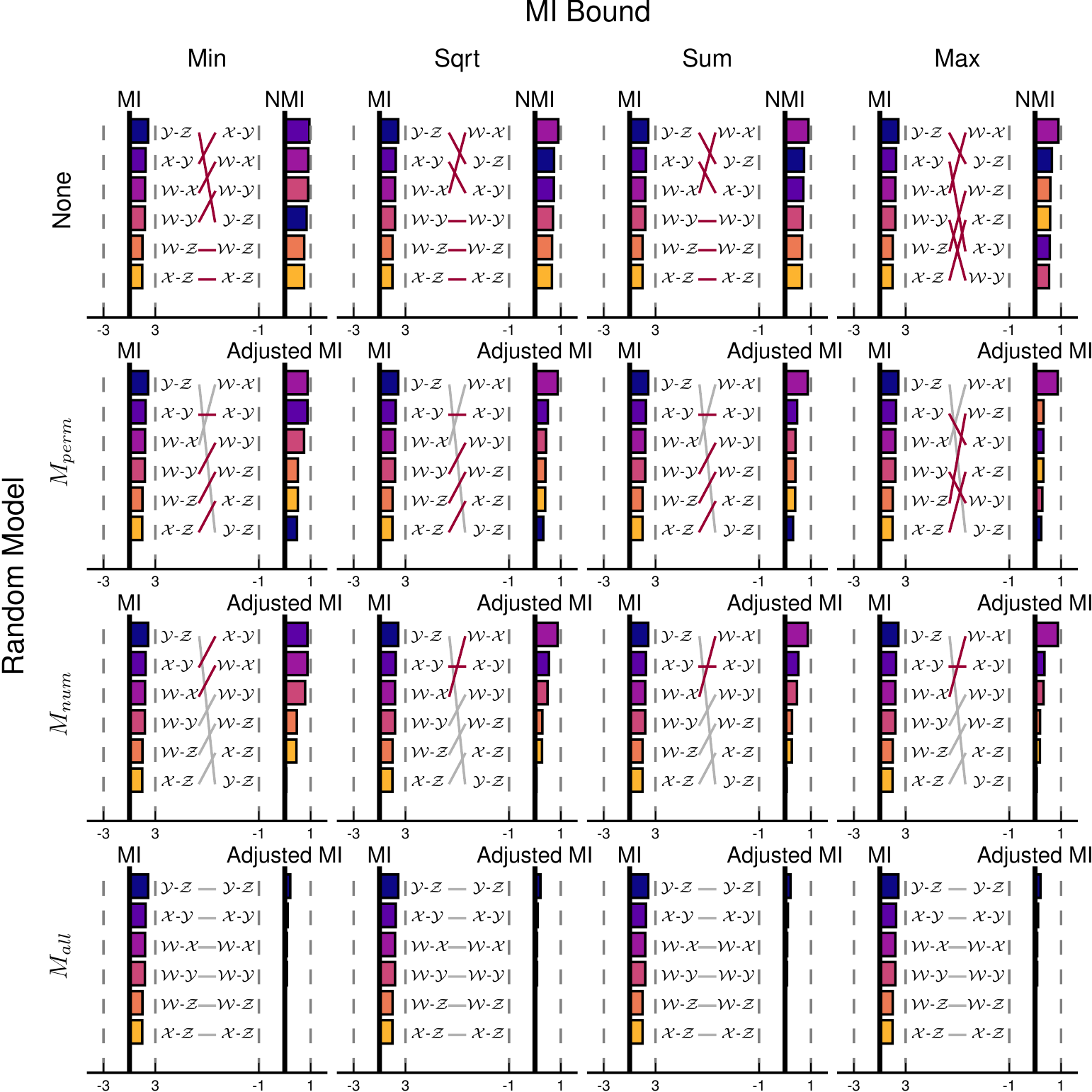}
	\caption{Both the random model and maximum bound have a significant impact on the rankings of clustering similarity using Mutual Information (MI). Rankings for the similarity of clustering pairs from \figureref{fig:randranking}a using (vertically) the raw MI, the Adjusted MI assuming $M_{\text{perm}}$, the Adjusted MI assuming $M_{\text{num}}$, and the Adjusted MI assuming $M_{\text{all}}$.  MI similarity also depends on the choice of maximum bound (horizontal) as a function of the two clusterings' model entropies; minimum (Min), square-root (Sqrt), average (Sum), and maximum (Max).  The MI measures which are normalized but not adjusted by a random model are all members of the common family of normalized MI (NMI).  For a given random model, similarity rankings which change as a function of maximum bound are highlighted in dark red.}
	\label{fig:miranking}
\end{figure}
\clearpage }

We then turn our attention to clustering similarity measured by mutual information (MI).
Rankings using the adjusted MI depend on two dimensions of variation: the random model and the maximum bound for the measure.  
This variation is illustrated in \figureref{fig:miranking} using the same 6 comparisons between pairs of clusterings from \figureref{fig:randranking}a.
We consider four cases for the MI maximum bound: Min, Sqrt, Sum, and Max, corresponding to the minimum of the two model partition entropies, the geometric mean of the two model partition entropies, the average of the two model partition entropies, and the maximum of the two model partition entropies, respectively (see \appendixref{sec:mi} for details).
For the permutation model, the model partition entropies are calculated from the cluster size sequences, while the model partition entropies in $M_{\text{num}}$ are bounded by the logarithm of the number of clusters and the model partition entropies in $M_{\text{all}}$ are bounded by the logarithm of the number of elements.
As a point of reference, all rankings are illustrated in comparison to the raw mutual information score, unnormalized and without a random model adjustment (None).
All adjustments of the mutual information without a random model (None, first row) are members of the commonly used family of Normalized Mutual Information (NMI) measures (\citealp{Danon2005communitycompare}).

The rankings in \figureref{fig:miranking} demonstrate that both the random model and the maximum bound affect the relative similarity between clusterings when adjusting MI.
Firstly, the only random model whose adjustments are independent of MI's maximum bound is $M_{\text{all}}$.  This occurs because every choice of the maximum bound reduces to $\log N$ (the entropy of the clustering that places each element into its own cluster).
In the other three random model scenarios, the maximum bound depends on the clusterings under comparison.
Secondly, MI is highly dependent on the number of clusters in each of the clusterings: When either no normalization and random model adjustment are used, or the $M_{\text{all}}$ model is used, MI ranks the similarity of clusterings $\mathcal{Y}$ and $\mathcal{Z}$ above that of $\mathcal{W}$ and $\mathcal{X}$  because of the greater number of clusters in the former case.
This bias is mitigated to varying extents by the NMI variations; while NMI using the Sqrt, Sum, and Max normalization terms all produce the intuitive ranking of $\mathcal{W}$ and $\mathcal{X}$ as the most similar pair, NMI using Min for normalization still succumbs to the larger number of clusters in $\mathcal{Y}$, and ranks $\mathcal{X}$ and $\mathcal{Y}$ as the most similar clustering pair.
The adjustments provided by both the $M_{\text{perm}}$ and $M_{\text{num}}$ random models control for the number of clusters; this reduces the impact of the number of clusters when the cluster sizes are regular, but the bias re-occurs when there is a large imbalance between the cluster sizes.

\subsection{Appropriate Random Model for Comparing K-means Clusterings}
\label{sec:kmeanexample}

Clustering similarity measures are commonly used to evaluate the results of clustering methods in relation to a known reference clustering.
Since the number of clusters can vary between instances, appropriately corrected similarity measures are necessary.
However, as we have already seen, the choice of similarity measure and its chance corrected variants can affect the results of the comparisons and suggest drastically different interpretations for the effectiveness of the method.

\afterpage{
\begin{figure}[ht]
	\centering
	\includegraphics[width = \textwidth]{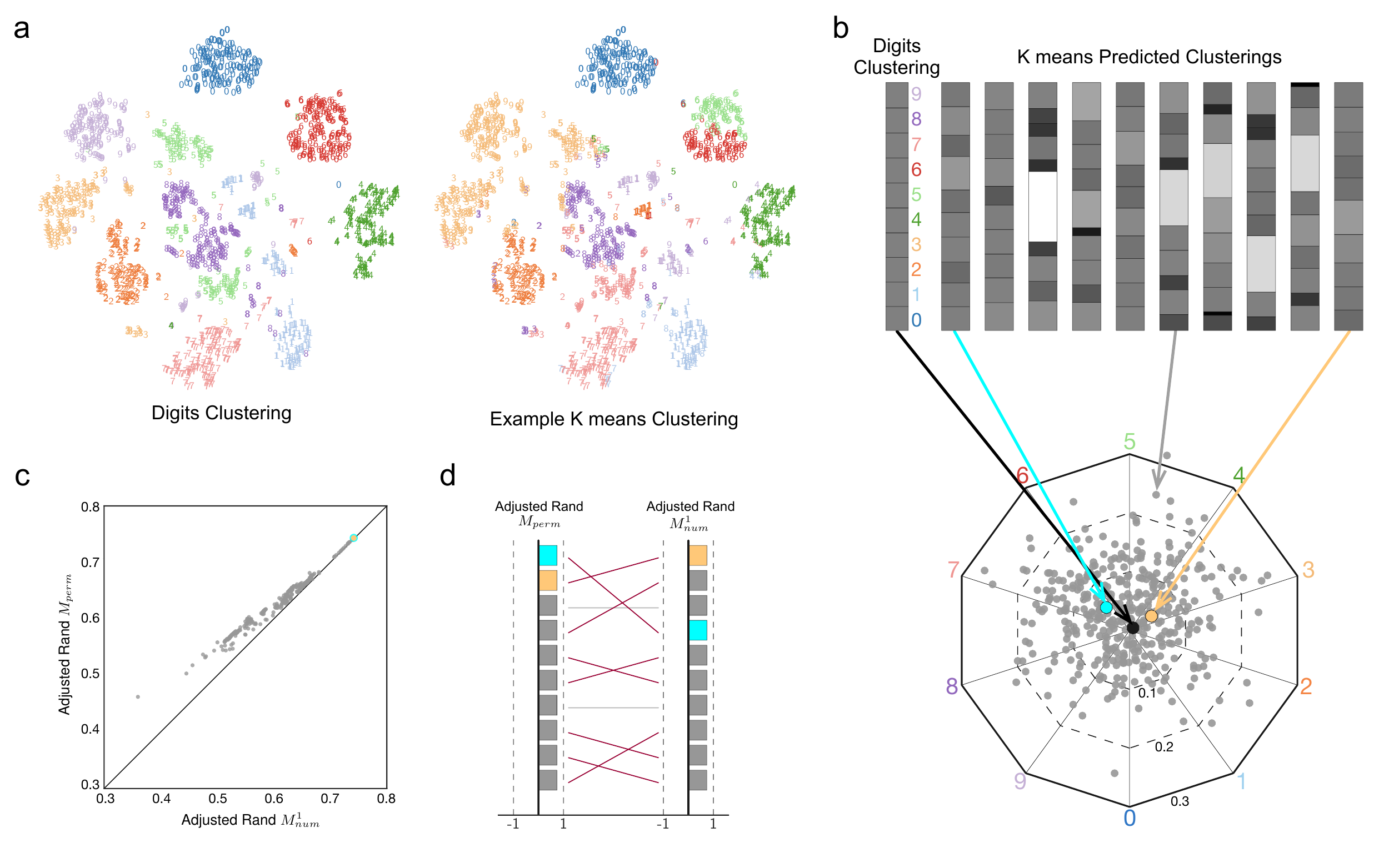}
	\caption{The impact of random model choice on the evaluation of K-means clustering.  \textbf{a}, The digits data set contains $1,797$ points in $64$ dimensions (projected to $2$ dimensions using t-SNE dimensionality reduction for visualization, \citealp{vanderMaaten2008tsne}) with a ground truth clustering corresponding to the digit, and an example K-means clustering.  \textbf{b}, The original cluster size sequence (top left) and $10$ cluster size sequences uncovered by K-means clustering with random initialization (top right). Intensity represents cluster sizes that are smaller (darker) or larger (lighter) than the ground truth clusters.  (bottom) The cluster size sequence for $400$ clusterings uncovered by K-means clustering with random initialization using Barycentric coordinates.  The actual clustering size sequence (black) and the most similar clustering determined by the Adjusted Rand index assuming $M_{\text{perm}}$ (light blue) and $M_{\text{num}}^1$ (light orange).  \textbf{c}, The similarity between the actual digits clusterings and each of the $400$ K-means clusterings as measured by the Adjusted Rand index assuming $M_{\text{perm}}$ (y-axis) and $M_{\text{num}}^1$ (x-axis).  \textbf{d}, The ranking of the most similar $10$ K-means clusterings as determined by the Adjusted Rand index assuming $M_{\text{perm}}$ (left) and $M_{\text{num}}^1$ (right).}
	\label{fig:digits}
\end{figure}
\clearpage}

We demonstrate the importance of the random ensemble assumption through a comparison of the clusterings uncovered by $400$ runs of K-means on a collection of hand-written digits (\citealp{Alimoglu1996digitsdata}, see \appendixref{sec:digitsdata} for details). 
The K-means clustering method groups elements so as to minimize the average (Euclidean) distance from the cluster centroid.  In most scenarios, it uncovers clusterings with a pre-specified number of clusters (K).
For our example, the digits naturally fall into $10$ disjoint clusters, shown in \figureref{fig:digits}a, with relative cluster sizes given on the left of \figureref{fig:digits}b.
Interestingly, almost all $400$ clusterings produced by K-means have a different cluster size sequence (\figureref{fig:digits}b, bottom) and the cluster sizes vary over a wide range (\figureref{fig:digits}b, top).   
This suggests that both the specific assignment of elements to clusters and the size sequence of the clusters are major factors differentiating the K-means clusterings.  Both sources of variation need to be captured by the random model in order to have a meaningful baseline.

\emph{Since the number of clusters does not change between runs, but the size sequence of those clusters changes considerably, it is more appropriate to assess similarity within the context of random clusterings with a fixed number of clusters rather than those given by the permutation model.}
Furthermore, since all of the comparisons are made against the same reference clustering, a one-sided similarity metric better captures the comparison scenario.
In \figureref{fig:digits}c, the similarity of the reference clustering compared to each of the $400$ uncovered clusterings is shown using the Adjusted Rand index assuming $M_{\text{perm}}$ and the Adjusted Rand index assuming $M_{\text{num}}^1$.  
While the measures are strongly correlated (the black line indicates perfect agreement), the Adjusted Rand index assuming $M_{\text{perm}}$ is consistently biased towards higher similarity.
Most importantly, the bulk of the uncovered clusterings change their relative ranking when considered in the context of $M_{\text{num}}^1$ compared to $M_{\text{perm}}$ as demonstrated by the rankings of the top $10$ most similar clusterings in \figureref{fig:digits}d.

\subsection{Random Models and Inhomogeneous Cluster Sizes}
\label{sec:paexample}

\emph{For both the Rand index and MI, the permutation model is invariant to differences in the cluster size sequence.}
This invariance is explicitly demonstrated in our next example by the difference between the adjusted similarity measures assuming $M_{\text{perm}}$ and $M_{\text{num}}$.
To generate an increasing disparity in cluster sizes, we use a preferential attachment model of element assignment.  At each step of the algorithm, a random element is uniformly chosen for reassignment to a new cluster based on the current sizes of those clusters.  A move is rejected if it results in an empty cluster.

\begin{figure}[th]
	\centering
	\includegraphics[width = \columnwidth]{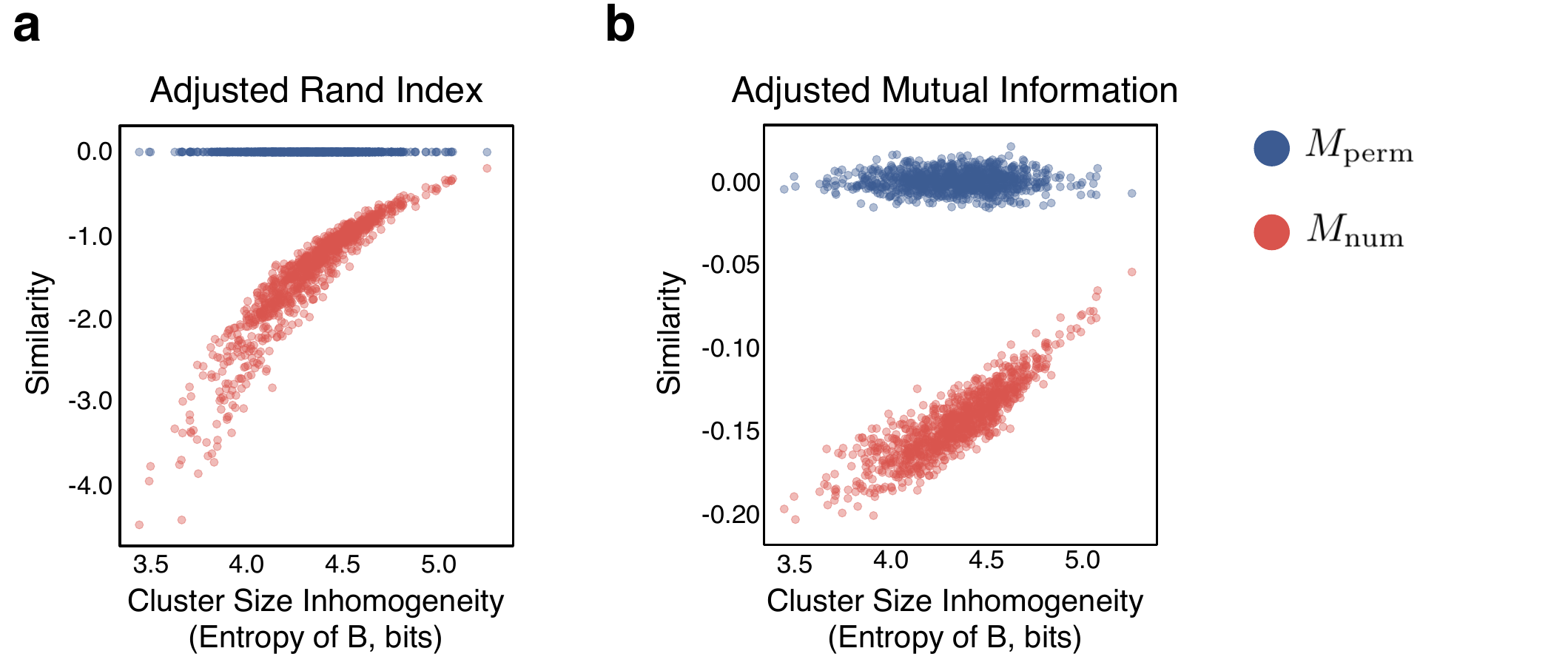}
	\caption{The invariance to inhomogeneous cluster size sequences when assuming $M_{\text{perm}}$.  A clustering with $50$ equal-sized clusters is compared to a second clustering $B$ generated by a preferential attachment model.  The cluster size sequence inhomogenity for clustering $B$ is measured by the cluster size sequence entropy (low entropy is indicative of large cluster size inhomogenity).  The similarity is calculated using the adjusted similarity assuming the permutation model $M_{\text{perm}}$ and $M_{\text{num}}$ for \textbf{a}, the Adjusted Rand index, and \textbf{b}, the Adjusted Mutual Information.  In both cases, the similarity assuming $M_{\text{perm}}$ is relatively constant (near 0), while the similarity assuming $M_{\text{num}}$ increases with increasing entropy.}
	\label{fig:pref}
\end{figure}

In \figureref{fig:pref}, we compare a clustering of $1,000$ elements grouped into $50$ equally sized clusters and a randomized variant of the same clustering throughout $10^6$-steps of our preferential attachment algorithm using the Adjusted Rand index (\figureref{fig:pref}a) and Adjusted MI (\figureref{fig:pref}b).
Cluster size inhomogeneity is measured by the entropy of the clustering size sequence; equally sized clusters have the maximum entropy ($\log_2 50 \approx 5.64$), while greater inhomogeneity in cluster sizes decreases the entropy of the cluster size sequence.
In both cases, the comparisons assuming $M_{\text{perm}}$ are invariant to the inhomogeneity of the cluster size sequences. 
On the other hand, comparisons assuming $M_{\text{num}}$ reflect the changes in the cluster size sequence.

\subsection{Performing at Random in Tumor Gene Expression Clustering}
\label{sec:geneexample}
Finally, recall that adjusted clustering similarity measures have the added interpretation with respect to a random baseline.
Such random baselines play an important role when evaluating methods in unsupervised learning and classification.
In our case, the adjusted similarity measures answer the question: \emph{Is the result of our clustering method more similar to the desired clustering than if we selected a random clustering?} 
The adjusted similarity measure quantifies an answer to this question: positive scores indicate performance above random, while negative scores indicate a random clustering is more similar.

\emph{The interpretation of the adjusted similarity as a random baseline is highly dependent on the assumption of the random model.}
Critically, if the random model does not reflect the actual ensemble in which the clustering method is searching, the baseline does not accurately reflect the scenario in question.
Thus, methods are incorrectly assessed as performing better than randomly generating a clustering.

\begin{figure}[t]
	\centering
	\includegraphics[width = \columnwidth]{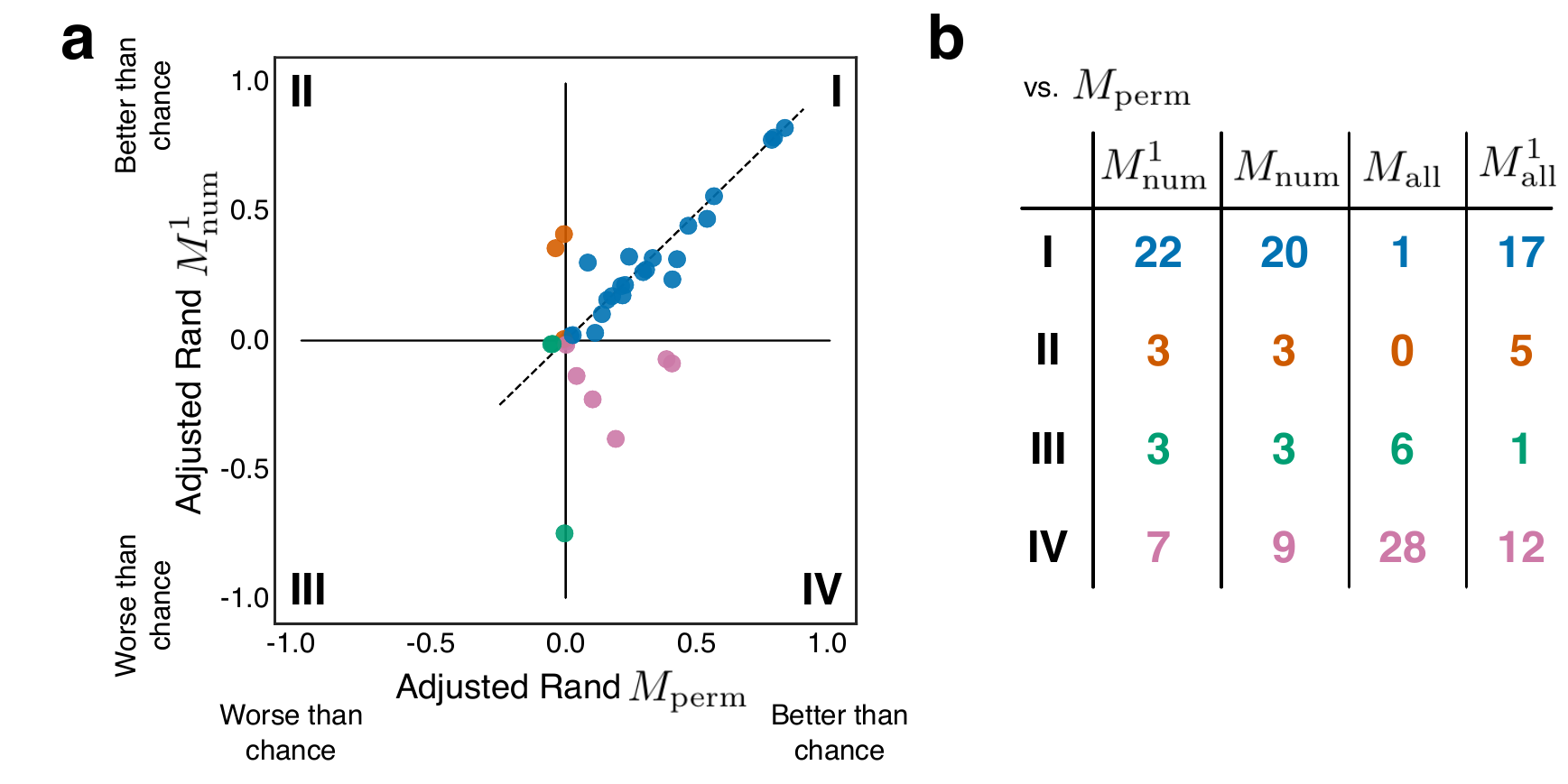}
	\caption{The impact of random model choice on the evaluation of gene expression clustering with respect to the random baseline.  The results of agglomerative hierarchical clustering identified from the gene expression in tissue samples from cancerous and healthy cells in $35$ studies.  \textbf{a}, The uncovered clusterings are compared to the reference clustering using the Adjusted Rand index assuming the permutation model $M_{\text{perm}}$ (x-axis), and the one-sided Adjusted Rand index assuming a fixed-number of clusters $M_{\text{num}}^1$ (y-axis).  The dashed grey line indicates numerical agreement between the similarity measures.  There are four possibilities when using two measures to assess similarity with respect to the random baseline: both random models conclude better than chance (blue, quadrant I), both random models conclude worse than chance (green, quadrant III), $M_{\text{perm}}$ concludes better than chance but $M_{\text{num}}^1$ concludes worse than chance (pink, quadrant IV), and visa-versa (orange, quadrant II).  \textbf{b}, The assumed random model affects the classification of clustering comparisons with respect to the random baseline in all four random models considered here ($M_{\text{num}}^1, M_{\text{num}}, M_{\text{all}}$ and $M_{\text{all}}^1$) vs.\ $M_{\text{perm}}$.}
	\label{fig:genes}
\end{figure}

We illustrate the dependence of adjusted similarity baseline on the choice of random model using a gene expression data set.
Specifically, we use a collection of $35$ cancer gene expression studies assembled in \cite{deSouto2008geneclustering}.
The studies in the collection aim to differentiate the gene expression in cancerous cell tissue samples from those in healthy controls.
Each study contains anywhere from $22$ to $248$ data points (individual tissue samples) for which between $85$ and $4,553$ features (individual gene expression) were measured after removing the uninformative and missing genes.
For details on the individual studies and filtering methodologies, see \cite{deSouto2008geneclustering} and references therein.

Clusterings are identified via agglomerative hierarchical clustering using correlation to compute the average linkage between data points, a common clustering methodology in biology.
While many other methods could be used (and indeed, were compared in \citealp{deSouto2008geneclustering}), we use hierarchical clustering as a representative example to illustrate the consequences of the random model.
Since hierarchical clustering produces a clustering with the user specified number of clusters, its similarity should be adjusted using the one-sided Adjusted Rand index assuming $M_{\text{num}}^1$, where the reference clustering is specified for each study individually.

\figureref{fig:genes} shows the similarity between the derived clustering and the reference clustering for each of the $35$ studies.
The Adjusted Rand index assuming $M_{\text{perm}}$ is shown on the x-axis; positive scores (blue and pink points) denote the method performed better than the random baseline, while negative scores (orange and green points) denote the method performed worse than the random baseline. 
When the Adjusted Rand index assuming $M_{\text{num}}^1$ is used (y-axis), a different classification of method performance with respect to the random baseline is found. 
Of particular note are the seven studies for which the method performed better than chance according to $M_{\text{perm}}$, yet, $M_{\text{num}}^1$ concludes the method actually performed worse than chance (pink points).
\emph{In this case, a random clustering drawn from the model with a fixed number of clusters would actually perform better than agglomerative hierarchical clustering, yet the practitioner using the permutation model would incorrectly conclude the method was performing better than chance.}
This discrepancy occurs even when the values of the Adjusted Rand index assuming $M_{\text{perm}}$ are relatively high ($>0.4$).
Similarly interesting are the three studies in which the method performed worse than chance according to $M_{\text{perm}}$, yet, $M_{\text{num}}^1$ concludes the method actually performed better than chance (orange points).

\section{Discussion}
Given the prevalence of clustering methods for analyzing data, clustering comparison is a fundamental problem that is pertinent to numerous areas of science. 
In particular, the correction of clustering similarity for chance serves to establish a baseline that facilitates comparisons between different clustering solutions.
Expanding previous studies on the selection of an appropriate model for random clusterings (\citealp{Meila2005axiomclusteringscompare, Vinh2009correctingnmi, Romano2016adjusting}), our work provides an extensive summary of random models and clearly demonstrates the strong impact of the random model on the interpretation of clustering results.

Our results underpin the importance of selecting the appropriate random model for a given context.
To that end, we offer the following guidelines:
\begin{enumerate}
\item Consider what is fixed by the clustering method: do all clusterings have a user specified number of clusters (use $M_{\text{num}}$), or is the cluster size sequence fixed (use $M_{\text{perm}}$)?
\item Is the comparison against a reference clustering (use a one-sided comparison), or are you comparing two derived clusterings (then use a two-sided comparison)?
\end{enumerate}

The specific comparisons studied here are not meant to establish the superiority of a particular clustering identification technique or a specific random clustering model, rather, they illustrate the importance of the \emph{choice} of the random model.
Crucially, conclusions based on corrected similarity measures can change depending on the random model for clusterings.
Therefore, previous studies which did promote methods based on evidence from corrected similarity measures should be re-evaluated in the context of the appropriate random model for clusterings (\citealp{Yeung2001genemodel, deSouto2008geneclustering, Yeung2001pcagene, Thalamuthu2006geneclusteringeval, Mcnicholas2010modelgenemixure}).

Throughout this work, we assumed a uniform probability of selecting a partition given a constraint on the types of partitions in the ensemble.
However, other probability distributions could be used which better model the clusterings encountered in practice.  
For example, instead of using a uniform distribution over the number of clusters, one could consider an inferred distribution for the number of clusters actually uncovered by a given method (e.g.\ affinity propagation).  
This is particularly relevant when considering $M_{\text{all}}$, an extreme case for random partitions.
Additionally, given that many systems exhibit clusterings with a heavy-tailed cluster size sequence, clusterings with such skewed cluster size distributions could be favored.
Changes to the prior probabilities would likely change the expectations of the clustering similarity measures.

The behavior of the Rand index and Mutual Information in the context of the random clustering models discussed here further reveals problems with both measures.
Specifically, the expected similarity of random clusterings increases as the number of elements grows.
Intuition would suggest the opposite; the similarity of two randomly selected clusterings should decrease as the number of elements increases because it is harder to match the element memberships to clusters between two random clusterings.
Instead, both MI and Rand are dominated by the fact that the expected number of clusters and cluster size distribution are converging with increasing $N$ (\citealp{Mansour2012setpartitions}).
Our analysis also illustrates the dependency on the normalization term for MI, which, combined with a previously established bias on the number of clusters, suggests more care should be taken when interpreting the results of MI clustering comparisons.
%

In conclusion, our framework for the correction of clustering similarity for chance allows for more conscious comparisons between clusterings.
%
%
The practitioner should always provide justification for their choice of random clustering model and treatment of one-sided comparisons. 


\acks{We thank Ian Wood, Santosh Manicka, James Bagrow, Sune Lehmann, Aaron Clauset, Randall D. Beer, and Luis M. Rocha for helpful discussions.  We also thank two anonymous reviewers for their wonderful insights and suggestions to improve the paper.  A package to computer the adjusted similarity measures is available on the author's github:\\ \url{https://github.com/ajgates42/clusim}}

\appendix
\section{Stirling and Bell Numbers}
The Stirling number of the second kind $S(n,k)$ gives the number of ways to partition a set of $n$ elements into $k$ clusters, where
\begin{equation}
	S(n,k) = \frac{1}{k!}\sum_{j=0}^k(-1)^{k-j}\binom{k}{j}j^n.
\end{equation}
There are several recurrence relations which also give $S(n,k)$, one of the most useful is the relation
\begin{align}
	S(0,0) &= 1 \quad\quad S(n, 0) = S(0, n) = 0 \\
	S(n+1, k) &= kS(n,k) + S(n, k-1). \nonumber
\end{align}
As $n\rightarrow\infty$, an asymptotic approximation to the Stirling numbers of the second kind for a fixed $k$ is given by $S(n,k)\approx\frac{k^n}{k!}$.

The Bell number $B_n$ is the total number of clusterings over a set with $n$ elements.  It is related the Stirling numbers of the second kind by the summation over $k$ for a fixed $n$, $B_n = \sum_{k=0}^nS(n,k)$.
There is also a useful recurrence relation for Bell numbers: $B_{n+1} = \sum_{k=0}^n\binom{n}{k}B_{k}$.
As $n\rightarrow\infty$, an asymptotic approximation to the ratio of the $n$-th and $(n+1)$-th Bell numbers is $\frac{B_n}{B_{n+1}}\approx\frac{\log n}{n}$.
See \cite{Mansour2012setpartitions} for an extended discussion of both the Stirling numbers of the second kind and the Bell numbers.

In practice, calculating the Bell numbers and Stirling numbers of the second kind from their recurrence relations can be computationally expensive.  
However, many efficient approximations and implementations are available  (\citealp{Temme1993stirling}; \citealp{Mansour2012setpartitions}).
Here, we make use of the mpmath arbitrary precision library for Python developed by \cite{Johansson2013mpmath}.
This library takes advantage of Dobi{\`{n}}ski's Formula to approximate the Bell numbers (\citealp{Dobinski1877summirung}; \citealp{Chen1994dobinski}).

\section{Application Data Sets}
\subsection{Digits Data Set}
\label{sec:digitsdata}
The digits data set is bundled with the sci-kit learn source code and consists of $1,797$ images of $8 \times 8$ gray level pixels of handwritten digits.  The reference clustering contains $10$ clusters corresponding to the true digit.  The data set was originally assembled in \cite{Alimoglu1996digitsdata}.
To provide a visualization, the data was projected to 2-d using the t-Distributed Stochastic Neighbor Embedding (t-SNE) dimensionality reduction method (\citealp{vanderMaaten2008tsne}) initialized from the pca decomposition.

\subsection{Gene Expression Data Set}
\label{sec:gene}
The data was assembled in \cite{deSouto2008geneclustering} and is freely available from
\\ \url{http://bioinformatics.rutgers.edu/Publications/deSouto2008c/index.html}.
The studies represent two prominent methods for determining gene expression in cell tissue samples from cancer tumors or healthy controls, Affymetrix microarrays and cDNA microarrays, which, respectively, measure the number of RNA copies found in the cell and the ratio of the number of copies vs a control sample.
Each study contains anywhere from $22$ to $248$ data points (individual tissue samples) for which between $85$ and $4,553$ features (individual gene expression) were measured after removing the uninformative and missing genes.
Please see \cite{deSouto2008geneclustering} for details of this selection process.
\bibliography{GatesAhn17refs}

\end{document}